\documentclass{ijuc}
\usepackage[pdftex]{graphics}
\usepackage{color}

\begin{document}

\title{How to deal with natural language interfaces in a geolocation context?}

\author{Mohammed-Amine Abchir\inst{1}\email{mohammed-amine.abchir@deveryware.com}
\and Isis Truck\inst{2} 
\and Anna Pappa\inst{2}
}

\institute{Deveryware, Paris, France
\and
University Paris 8, Saint-Denis, France
}

\def\received{Received 1 December 2012; In final form xx yy zzzz}

\maketitle

\begin{abstract}
In the geolocation field where high-level programs and low-level devices coexist, it is often difficult to find a friendly user interface to configure all the parameters.
The challenge addressed in this paper is to propose intuitive and simple, thus natural language interfaces
to interact with low-level devices. Such interfaces contain natural language processing and fuzzy representations of words that facilitate the elicitation of business-level objectives in our context.

\end{abstract}


\section{Introduction}

The question of geolocation is the core base of many companies working on geomarketing. Social networks with recommendations (nearby social events, nearby restaurants, etc.), determination of a customer profile depending on his choices, preferences, tastes, etc. are examples of growth sectors, especially with the expansion of the mobile market that is rapidly growing. The company we are working with, is a leader in geolocation middleware and proposes systems for child location, vehicle and fleet tracking or
delivery rounds. From the company point of view, persons, vehicles, goods are considered as devices that have to be tracked
with accuracy using a \emph{Geohub} and notifications must be sent according to type of required tracking. On the contrary, from the customer point of view, a service
has to be proposed, according to a prior agreement but no (or few) technical detail(s) has (have) to be known.
The Geohub coordinates geoinformation on a single platform to track devices and to give them orders knowing their position (\emph{e.g.} what to do in case of traffic jam, or accident on the road, or if the vehicle has broken down, etc.).
The Geohub interacts with all the tracked devices. Low-level messages (written in the Forth language) are sent by both
the devices and the Geohub in ``push'' or ``pull'' mode. 
Two roles have to be distinguished: (1) the company who sells and connects devices to its hub and who configures the hub to satisfy the customer and (2) the customer who expresses his need (his fleet tracking, for example) to an employee of the company that is able to configure the hub. Thus the employee has to understand the customer needs and elicit his preferences while configuring the Geohub.

An important challenge for the company is to propose a smart interface, smart enough to let the customers configure the Geohub by themselves. The customers would have a phone conversation with a virtual assistant. We know that
natural language processing (NLP) deals with NP-complete problems that is why such an interaction needs many contextual elements to limit the possibilities. However, even with context, the problem remains quite hard since it implies an important expertise to be able to translate the needs expressed in a natural language into a set of Forth scripts and programs written in other languages. Thus there is a need for tools to capture these sometimes imprecise requirements (\emph{e.g.} ``I would like to be notified when one of my fleet vehicles arrives near the warehouse'') and transform them into business processes. The fuzzy logic and its various methods and tools are of great help for such needs and especially the methods that deal with linguistic variables.

The paper is organized as follows: we first review some NLP techniques and we focus on an interesting linguistic fuzzy model to deal with imprecisions. Our approach that mixes NLP with fuzzy tools in the geolocation context is then explained. Finally, a use case is presented: it exhibits the use of a fuzzy semantic approach to activate an alert in a geolocation context.

\section{State of the art}
Since the first techniques to deal with natural language, many methods and algorithms have been proposed to understand
and disambiguate sentences~\cite{Cho86,Kuh90,Win72,Wei66}. For example, decision trees and statistical models may
give good results as soon as there is context enough in the speech corpus.
In NLP, a formal grammar draws a set of normative rules to describe how the natural language runs.
Among the various formal grammars are the generative grammars, the transformational grammars and the tree-adjoining grammar (TAG)~\cite{Jos83}.  To make discourse analysis, part-of-speech tagging is often used to disambiguate words
(\emph{e.g.} ``display'' can be a noun, a transitive verb or an intransitive verb)~\cite{Har91}. 
Syntax is part of formal grammar that deals with rules for the structure of a string (distribution of words, noun and adjective agreements, etc.).
Meaning and formal syntax are the basis of the comprehension, according to Chomsky~\cite{Cho75}. Semantics is the study of the meaning of words or sentences in their context. The concept of meaning is often fuzzy because natural language may be imprecise if we consider that each word (or set of words) has more than one meanings (without taking into account the figurative sense that is something else again). In linguistics, the meaning of a vague or imprecise expression is a fuzzy set in the proper sense. What we call \emph{fuzzy semantics} is an approach that uses fuzzy logic (Zadeh's sense) to express the fuzziness of the meaning.

Zadeh was the pioneer researcher to propose approaches to deal with imprecision and uncertainty when he introduced the fuzzy set theory, the fuzzy logic and the concept of linguistic variables~\cite{Zad65}. Since, many fuzzy models have been proposed for \emph{computing with words} but one seems the most appropriate in our case
because it makes a simple correspondence between words and fuzzy scales: the 2-tuple fuzzy linguistic model~\cite{Her00}. Each word or linguistic expression is translated into a linguistic pair $(s_i,\alpha)$ where $s_i$ is a triangular-shaped fuzzy set and $\alpha$ a symbolic translation. If $\alpha$ is positive then $s_i$ is reinforced else $s_i$ is weakened. If the information is perfectly balanced (\emph{i.e.} the distance between each consecutive word is exactly the same), then all the $s_i$ values are equally distributed on the axis. But if not (which may happen when talking about distance, for instance, ``almost arrived'' and ``close to'' are closer to each other than ``far'' and ``out of route'') the $s_i$ values may not be equally distributed on the axis. Another model has thus been proposed to deal with such information called multigranular linguistic information~\cite{Her08}. To perform the computations, linguistic hierarchies composed of an odd number of triangular fuzzy sets of the same shape, equally distributed on the axis, are used as a fuzzy partition.  A word may have one or two linguistic hierarchy(ies) and the representation of a notion (such as ``distance'') can be made up of several hierarchies. Thus the fuzzy sets obtained are still triangular-shaped but not always isosceles triangular-shaped and the notation $s_i^j$ permits to keep both the hierarchy and the linguistic term. See~\cite{Mar10} for a deeper review of these models.

Nevertheless, we have shown in recent papers that the 2-tuple model with unbalanced linguistic term sets don't allow to
model any unbalanced scale. Indeed, when one linguistic expression is very far away from its next neighbor,
the 2-tuple fuzzy linguistic model puts them artificially closer to eachother~\cite{Abc11a,Abc11b}. The model we have proposed that is of course inspired by the 2-tuple fuzzy linguistic model uses the symbolic translations $\alpha$ to generate the data set.
In our model, the 2-tuples are twofold. Except the first one and the last one of the partition, they all are composed of two half 2-tuples: an upside and a downside 2-tuple. In our context (geolocation) where the linguistic terms are usually unbalanced, this choice is quite relevant.

\section{How a linguistic 2-tuple model can help NLP?}

Parts of Speech (PoS) recognition and the PoS tagging have been used to analyze the strings. The analysis is simplified using a semantic tagging because the context (geolocation software) is known. As an example, let us take the following need:
``\textsf{I want to receive an alert when the vehicle gets very close to the warehouse}''. The tagging permits to obtain \emph{tokens}
with a certain grammar and meaning. ``\textsf{I}'' is grammatically a pronoun and has no particular meaning for us (it simply 
designates the customer). ``\textsf{alert}'' is grammatically a noun and has meaning \verb£ALERT£. ``\textsf{gets}'' is grammatically a verb and has meaning \verb£ZONE_ENTRY£. ``\textsf{very}'' is grammatically an adverb and has meaning \verb£FUZZY_MODIF_+£. ``\textsf{close to}'' is grammatically an adjective and has meaning \verb£DISTANCE£.

A tree using a simplified tree-adjoining grammar (TAG)-based is then created, where the root is an \verb£ALERT£.
Each leaf node represents the semantic tag of a token from the lexicon. In the example below \verb£POI£ means ``point of interest'' and \verb£ZOI£ ``zone of interest''.
\vspace*{-.5cm}
\begin{verbatim}

ALERT=TYPE,MOBILE,PLACE,NOTIFICATION
TYPE=ZONE_ENTRY|ZONE_EXIT|CORRIDOR
...
PLACE=TOWN|ADDRESS|POI|ZOI
...
\end{verbatim}

\begin{figure}[!h]
  \begin{center}
    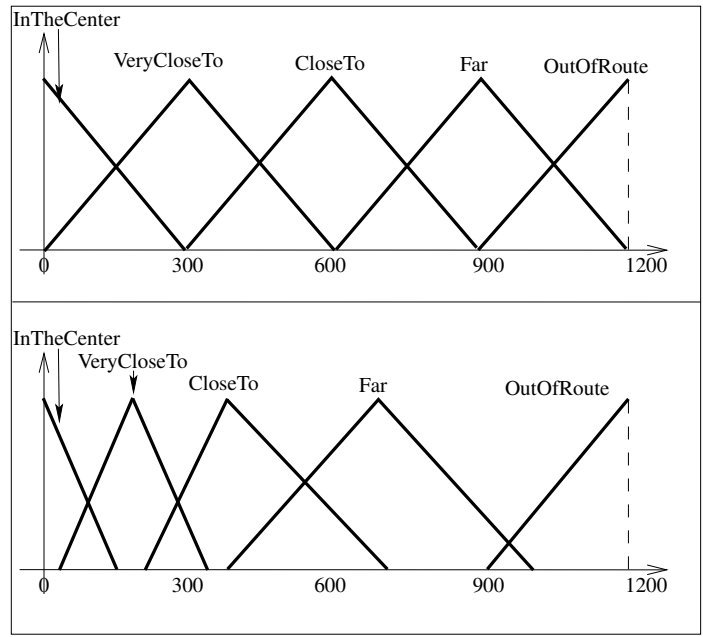
    \caption{\label{fig:2tuples}Unbalanced linguistic term sets: example of our partition model for distance.}
  \end{center}
\end{figure}

For each concept whose meaning may be imprecise or fuzzy, a fuzzy partition with our 2-tuple model is performed by experts. As an example, if the expert gives five labels to represent the concept \verb£DISTANCE£, they are by default considered as uniformly distributed on their axis. Looking for synonyms (\emph{e.g.} \verb£http://www.crisco.unicaen.fr/cgi-bin/cherches.cgi£ that exhibits a French dictionary) gives a set of five synonym bags, one bag per label.
The distance between each label is computed using the number of shared synonyms in each bag. Resemblance rate matrices are then computed to determine both the \emph{order} of the terms and the \emph{distance} between them.
Finally it is easy to construct the partition of the five unbalanced terms thanks to our 2-tuple linguistic model. See Figure~\ref{fig:2tuples} where both partitions (top, Herrera \& Mart\'inez's one and down, ours) are shown.

At the top of the figure,
the partition seems perfectly balanced because the model forces to have a midterm (here, ``CloseTo''). The rest of the terms
are automatically placed from left to right. At the bottom of the figure, our partition is unbalanced because we
used our method with semantics and synonyms. Of course the big difference between both models is that in our case fuzzy subset
``Far'' meets ``OutOfRoute'' with a membership degree lesser than .5 (.15 actually). But our model guarantees the minimal coverage property, as proved in~\cite{Abc13}, that is enough in the computations.

The semantic tokens (\emph{e.g.} ``\textsf{close to}'') are then expressed through our 2-tuples and compared to the fuzzy partition. Adverbs such as ``\textsf{very}'' will modify their associated 2-tuple through the $\alpha$ translation value, that has to be seen as a \emph{modifier}.
We have implemented on Android platform a vocal interface application built with PhoneGap, an open-source development framework that uses the phone's vocal recognition API.

In the following we propose a use case using both fuzzy models: Herrera \& Mart\'inez' one and ours.

\section{Use case and comparison of the fuzzy models}

As a use case we keep the same example than before with an alert that is triggered depending on (1) the distance between the mobile and the arrival (a warehouse) that is subject to (2) a time tolerance and depending on (3) the battery level.
If the time tolerance is high enough, the alert is not triggered when the mobile is quite far away from the arrival and the battery level is low (to save battery). The alert is triggered, even if the battery level is low, when the mobile drives out-of-route of the arrival. The software has been written in Java, using jFuzzyLogic with the extension we have proposed~\cite{Abc11b} and the FCL (Fuzzy Control Language) specification (IEC 61131-7). 
The FCL script below uses three inputs and one output.

{\scriptsize \begin{verbatim}
FUNCTION_BLOCK Alert1
VAR_INPUT	Battery:LING; Distance:LING; TimeTolerance:LING;
END_VAR
VAR_OUTPUT
        AlertTrigger:REAL;
END_VAR
FUZZIFY Battery
        TERM S := pairs (Minimum, 0.0) (VeryLow, 10.0) (Low, 20.0)
                        (Medium, 50.0) (High, 60.0) (VeryHigh, 80.0)
                        (Maximum, 100.0);
END_FUZZIFY
FUZZIFY Distance
        TERM S := pairs (InTheCenter,0.0) (VeryCloseTo,200.0) (Near,400.0)
                        (Far,700.0) (OutOfRoute,1200.0);
END_FUZZIFY

FUZZIFY TimeTolerance
        TERM S := pairs (Minimum, 0.0) (Medium, 60.0) (Maximum, 120.0);
END_FUZZIFY

DEFUZZIFY AlertTrigger
        TERM NoAlert := trian 0.0 0.0 1.0;
        TERM Alert := trian 0.0 1.0 1.0;
        METHOD : COG;		// 'Center Of Gravity'
END_DEFUZZIFY

RULEBLOCK Rules
        RULE 1: IF Battery IS Minimum AND Distance IS InTheCenter AND
	                   TimeTolerance IS Minimum THEN AlertTrigger IS NoAlert;
        ...                   
        RULE 28: IF Battery IS Maximum AND Distance IS Far AND
	                    TimeTolerance IS Minimum THEN AlertTrigger IS Alert;
        ...
        RULE 63: IF Battery IS Maximum AND Distance IS Far AND
	                    TimeTolerance IS Medium THEN AlertTrigger IS Alert;
        ...
END_RULEBLOCK

END_FUNCTION_BLOCK	
\end{verbatim}}

Two different partitions were used for the inputs: the 2-tuple partition from Herrera \emph{et al.} and our 2-tuple partition (in the FCL extract, only our partition is shown through the \verb£pairs£).
A study and a comparison between both models show that no alert is triggered with Herrera \emph{et al.}'s model when
(i) distance is between \verb£Near£ and \verb£Far£ (=600) and battery level is \verb£Maximum£ (=100); (ii) distance is \verb£Far£ (=700) and battery level is \verb£Maximum£ while an alert would have been triggered with our 2-tuple model.
 This is due to the fact that the unbalancement is weaker in this model than in ours.
 


\section{Conclusion}

This paper is a first approach to take into account natural language in a geolocation interface where an implementation has been proposed on Android platform. The imprecision, \emph{i.e.} the fuzzy semantics of the sentences are dealt with a model inspired by the 2-tuple fuzzy linguistic representation model from Herrera \& Mart\'inez where unbalancement is the key concept. The partition is constructed while taking into account the semantics of the concepts (synonyms, resemblance rate matrices, etc.). In future works, we will formalize this method that is empirical so far. We also want to introduce more contextual elements and ontologies to be able to understand more words.

\section{Acknowledgments}

This work is partially supported by the French National Research Agency (ANR) under grant number ANR-09-SEGI-012.

\bibliography{ATP}


\end{document}